\documentclass[10pt,twocolumn,letterpaper]{article}

\usepackage{iccv}              
\definecolor{iccvblue}{rgb}{0.21,0.49,0.74}
\usepackage[pagebackref,breaklinks,colorlinks,allcolors=iccvblue]{hyperref}

\title{Enhancing Domain Diversity in Synthetic Data Face Recognition with Dataset Fusion}

\author{Anjith George and  Sébastien Marcel \\
Idiap Research Institute, Switzerland\\
{\tt\small \{anjith.george,sebastien.marcel\}@idiap.ch}\\}

\begin{document}
\maketitle
\begin{abstract}
While the accuracy of face recognition systems has improved significantly in recent years, the datasets used to train these models are often collected through web crawling without the explicit consent of users, raising ethical and privacy concerns. To address this, many recent approaches have explored the use of synthetic data for training face recognition models. However, these models typically underperform compared to those trained on real-world data. A common limitation is that a single generator model is often used to create the entire synthetic dataset, leading to model-specific artifacts that may cause overfitting to the generator’s inherent biases and artifacts. In this work, we propose a solution by combining two state-of-the-art synthetic face datasets generated using architecturally distinct backbones. This fusion reduces model-specific artifacts, enhances diversity in pose, lighting, and demographics, and implicitly regularizes the face recognition model by emphasizing identity-relevant features. We evaluate the performance of models trained on this combined dataset using standard face recognition benchmarks and demonstrate that our approach achieves superior performance across many of these benchmarks.
\end{abstract}    
\section{Introduction}
\label{sec:intro}

\begin{figure*}[!h] 
               \centering
               \includegraphics[width=0.99\textwidth]{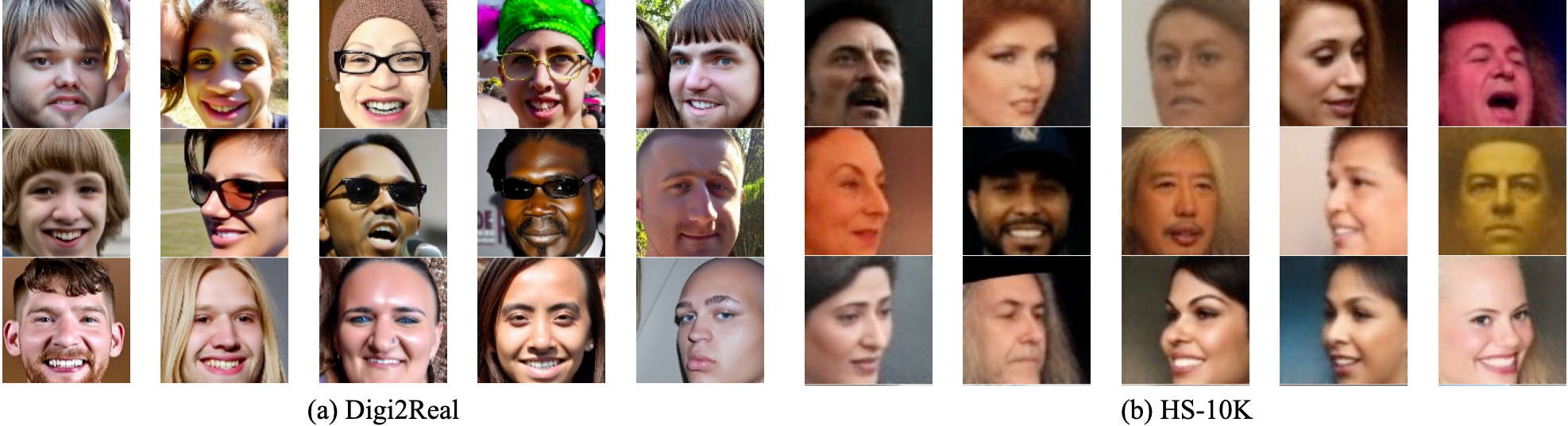}
               \caption{Images from (a) Digi2Real dataset \cite{george2025digi2real} and HS-10K \cite{wu2024vec2face}, showing different identities in these datasets. It can be seen that the distribution of images are very different in two datasets }
               \label{fig:dataset_interclass} 
\end{figure*}

Face recognition technology has become a widely adopted method for biometric authentication, largely driven by advances in deep neural networks and the availability of large-scale training datasets \cite{deng2019arcface, kim2022adaface, zhu2021webface260m, guo2016ms}. However, the use of these datasets has raised significant privacy concerns, particularly because many of them were compiled without obtaining informed consent from the individuals depicted. This practice poses serious legal and ethical challenges, especially in the context of data protection regulations such as the European Union's General Data Protection Regulation (GDPR). As a result, several legacy datasets have been withdrawn from public access. In response, there is a growing interest in using synthetic data for training face recognition models. This shift is reflected in the emergence of public benchmarks and competitions focused on synthetic face recognition datasets \cite{melzi2024frcsyn, otroshi2024sdfr, deandres2024frcsyn, melzi2024frcsynongoing, deandres2025second}.

Most existing efforts to generate synthetic face datasets rely on generative models such as StyleGAN \cite{karras2019style}, diffusion models \cite{rombach2022high}, or graphics-based rendering pipelines \cite{bae2023digiface}. While generative models can produce high-quality images, they typically require large amounts of real data to train the generator networks and often use datasets like FFHQ \cite{karras2019style} to learn the underlying face distribution. In contrast, graphics-based approaches, such as DigiFace-1M \cite{bae2023digiface} utilize rendering pipelines to synthesize face images without the need for extensive real-image datasets or pretrained face recognition networks. These methods draw on techniques similar to those in \cite{wood2021fake}, combining 3D facial geometry, textures, and hairstyles. This enables the generation of intra-class variations by altering pose, facial expression, illumination, and accessories. Notably, such pipelines offer the capability to synthesize a large number of unique identities with diverse intra-class variability and broad ethnic representation. Additionally, they support controlled generation by allowing specific attributes to be explicitly defined during synthesis.

Despite these advancements, models trained exclusively on synthetic datasets generally underperform compared to those trained on real-world data. A common limitation in existing synthetic datasets is that they are often generated entirely using a single generative model or pipeline. This can introduce generator-specific artifacts and limit the diversity of the synthesized data, ultimately affecting the model's generalization and recognition performance. 

In this work, we investigate the effectiveness of training face recognition models on a combination of two distinct synthetic datasets to enhance diversity and mitigate generator-specific biases. We also present our submission to the DataCV ICCV Challenge, which focuses on generating large-scale training datasets for face recognition. The datasets used and the protocol will be made available publicly \footnote{\url{https://gitlab.idiap.ch/biometric/code.datacv2025}}.

\section{Related works}
\label{sec:relworks}
Recent works in literature have explored synthetic dataset generation as a way to address the legal and ethical limitations of using real face data. Most approaches rely on generative models such as StyleGAN and Diffusion models. In the following section, we briefly review few of the prominent synthetic face datasets.

SynFace \cite{qiu2021synface} utilized DiscoFaceGAN \cite{deng2020disentangled} to study intra-class variance and domain gaps in synthetic face data. The method was enhanced with the identity and domain mixup method to increase diversity. SFace \cite{boutros2022sface} introduced a class-conditional GAN for labeled face synthesis, enabling supervised training of facial recognition (FR) models with competitive accuracy. IDiff-Face \cite{boutros2023idiff} employed a two-stage pipeline: first, an autoencoder was trained, followed by a conditional latent diffusion model. Conditioning was achieved via low-dimensional projections and cross-attention, with dropout preventing overfitting to identity features. GANDiffFace \cite{melzi2023gandiffface} combined StyleGAN with diffusion models. Initial demographic-aware identity generation using StyleGAN was followed by fine-tuning diffusion models to produce demographically diverse samples. ExFaceGAN \cite{boutros2023exfacegan} disentangled identity features within StyleGAN’s latent space by learning identity decision boundaries, allowing identity-consistent and diverse image synthesis. IDNet \cite{kolf2023identity} introduced a three-player GAN involving a pretrained FR network to guide StyleGAN toward generating identity-separable images, with only the classifier layers being updated.

SynthDistill \cite{shahreza2023synthdistill,shahreza2024knowledge} used feedback-based data generation to iteratively train lightweight models on challenging samples, which can be seen as a direct way of using generated images for the task of distillation. DisCo \cite{geissbuhler2024synthetic} proposed a Brownian-motion-inspired latent space sampling technique, combining identity dispersion and latent augmentation to generate diverse and identity-consistent samples.

While generative methods struggle to produce a large number of distinct, identity-consistent samples, DigiFace \cite{bae2023digiface} adopted a rendering-based approach using 511 consented 3D face scans. A parametric face model was constructed to generate over 1.2 million images with 110,000 unique identities by varying geometry, texture, hairstyle, and environment. Despite its scalability and controllability, the resulting images suffer from reduced realism, introducing a domain gap in FR model training. Rahimi et al. \cite{rahimi2024synthetic} applied image-to-image translation to enhance DigiFace images using pretrained models like CodeFormer. Their method improved visual quality without real identity-labeled data, though performance remained below models trained on real images.

Most existing methods show significantly lower performance compared to models trained on real data. This performance gap remains a key barrier to the practical adoption of synthetic data for face recognition.

\section{Proposed Method}

\begin{figure}[!h] 
               \centering
               \includegraphics[width=0.99\columnwidth]{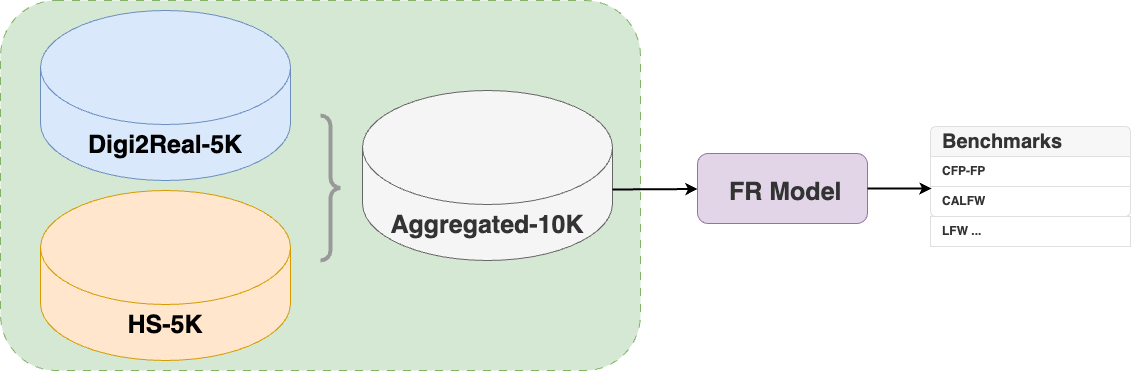}
               \caption{Aggregation of two datasets for training the models}
               \label{fig:framework} 
\end{figure}

As previously noted, one key limitation of many synthetic face datasets is that they are generated using a single model, typically a GAN or a diffusion-based generator. This can lead to models trained on such data overfitting to generator-specific artifacts and failing to generalize. To mitigate this issue, the primary contribution of this work is to combine datasets generated from different sources during training, thereby reducing dataset-specific biases. In our approach, we leverage two state-of-the-art synthetic face recognition datasets: Digi2Real \cite{george2025digi2real} and Vec2Face \cite{wu2024vec2face}. Samples from these datasets are shown in Figure \ref{fig:dataset_interclass}. These datasets were selected due to their strong performance on face recognition benchmarks and because they are generated using entirely different pipelines. The use of distinct generation pipelines contributes to the diversity of the combined dataset. The following subsection provides detailed information about these datasets.

\begin{figure}[!h] 
               \centering
               \includegraphics[width=0.95\columnwidth]{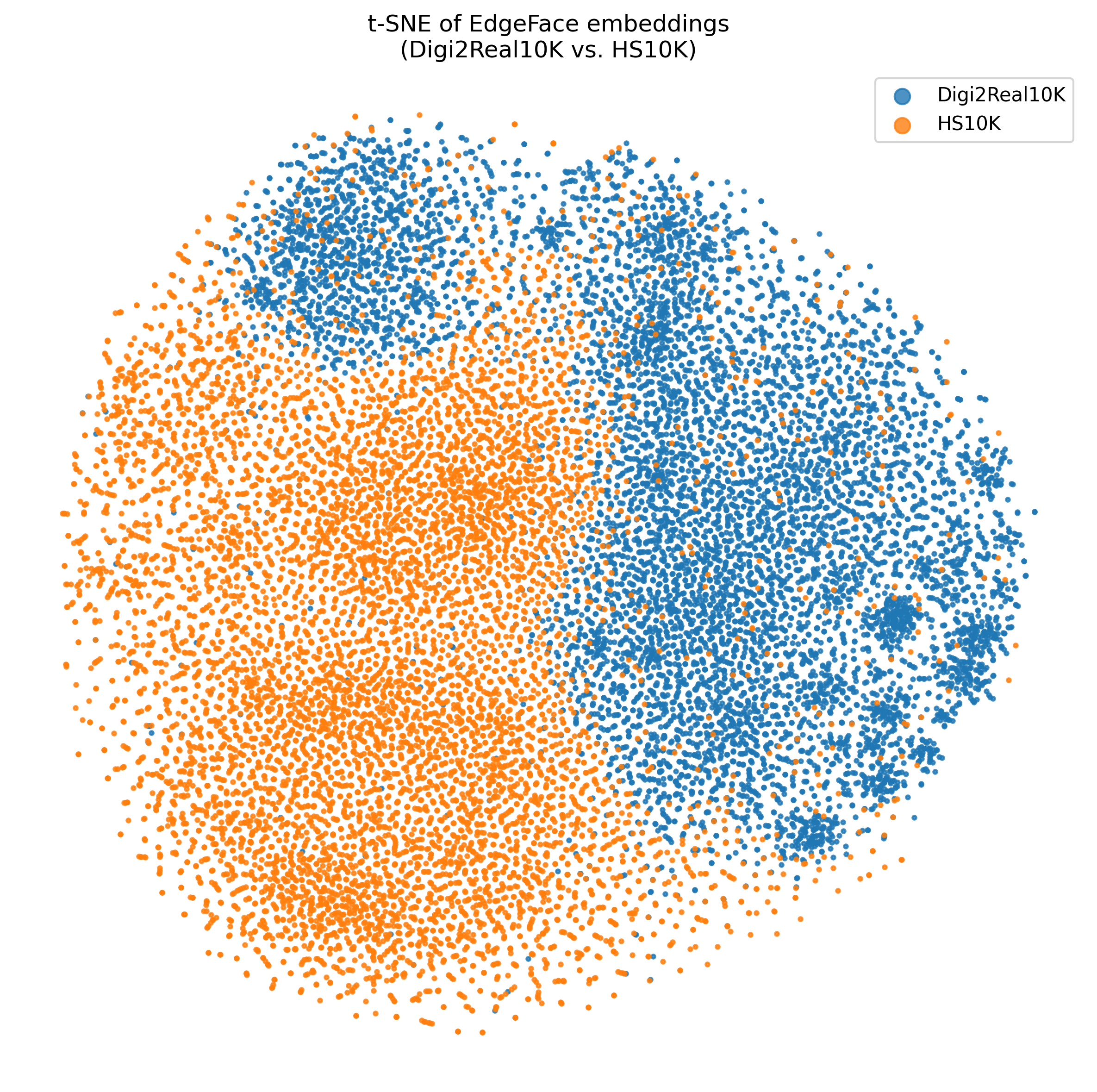}
               \caption{T-SNE plot showing the distribution of Digi2Real dataset \cite{george2025digi2real} and HS-10K \cite{wu2024vec2face} on the EdgeFace identity latent space.  }
               \label{fig:edgeface} 
\end{figure}

\begin{figure}[!h] 
               \centering
               \includegraphics[width=0.95\columnwidth]{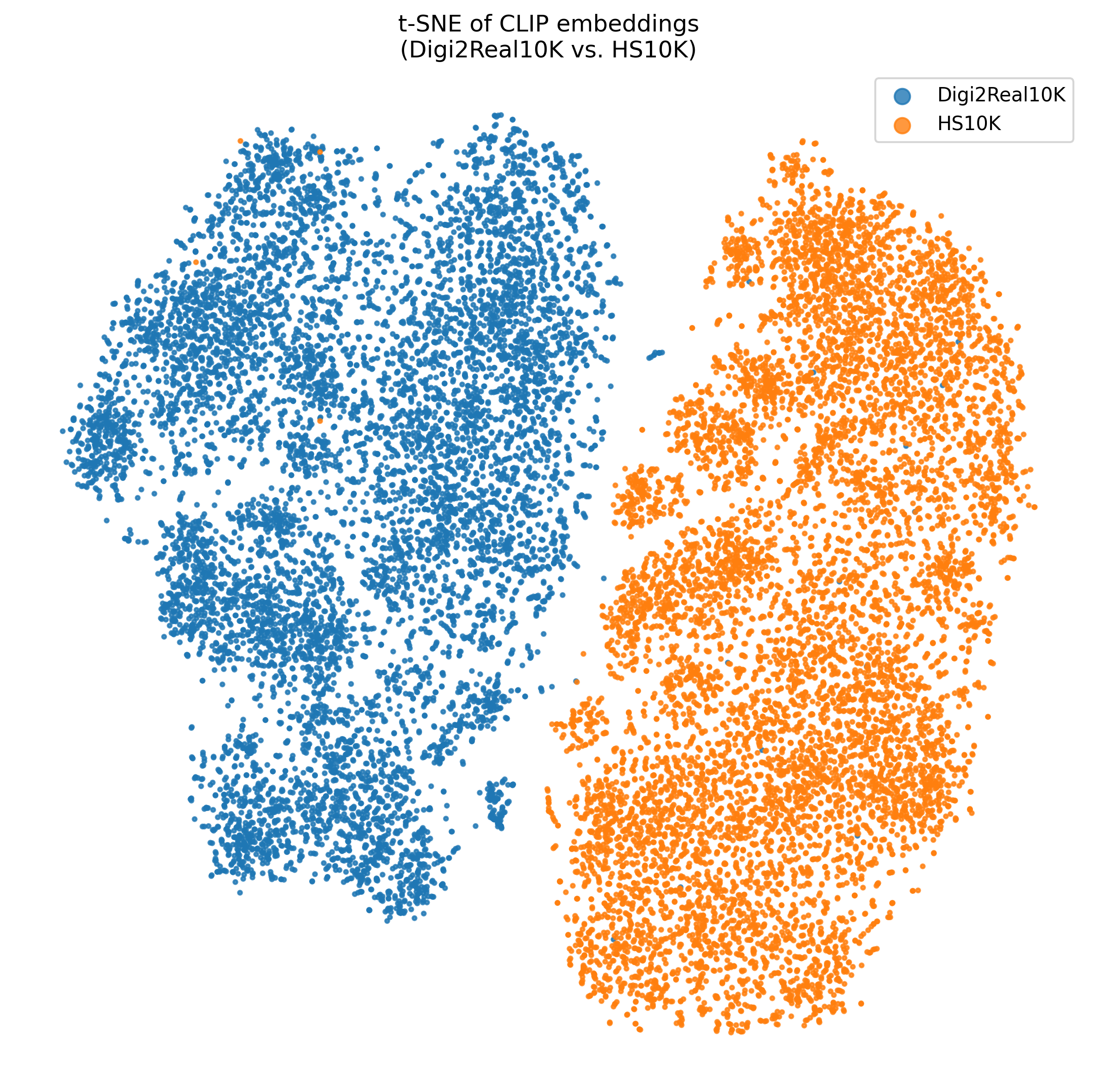}
               \caption{T-SNE plot showing the distribution of Digi2Real dataset \cite{george2025digi2real} and HS-10K \cite{wu2024vec2face} on the CLIP latent space. }
               \label{fig:clip} 
\end{figure}
\subsection{Digi2Real} Digi2Real \cite{george2025digi2real} introduces a hybrid approach that combines graphics-based rendering with generative methods to improve the realism and effectiveness of synthetic face datasets. Specifically, the method reuses identities from the DigiFace dataset and applies identity-preserving realism transfer to mitigate the domain gap between synthetic and real images. DigiFace images, while diverse, often appear cartoonish and suffer from a substantial domain gap, which negatively impacts the performance of models trained on them. Digi2Real addresses this by transforming DigiFace images into more realistic counterparts while preserving identity information. The pipeline also introduces new intra-class variations conditioned on the original identity features, using an identity latent space induced by the ArcFace model \cite{papantoniou2024arc2face}. It leverages the Arc2Face model \cite{papantoniou2024arc2face}, a face foundation model capable of generating face images conditioned on ArcFace embeddings. To enhance inter-class diversity, the method performs spherical linear interpolation (SLERP) in the latent space, while compensating for distribution shifts in the intermediate CLIP space to further reduce the domain gap. Final images are post-processed using face detection and cropping to produce a clean dataset. Experimental results demonstrate that Digi2Real significantly outperforms the original DigiFace dataset in face recognition tasks.

\subsection{Vec2Face} Vec2Face \cite{wu2024vec2face} presents a method for generating a large number of synthetic identities with controllable intra-class variations, enhanced through external attribute models. The architecture consists of a pretrained face recognition model, a feature-masked autoencoder, an image decoder, and a patch-based discriminator. After training, identity vectors are sampled using a multivariate normal distribution in the PCA-transformed latent space and then projected back to the original space. To ensure identity diversity, vectors that are too similar to existing ones are filtered out based on a similarity threshold. Intra-class variations are introduced through random perturbations of the identity vector and further refined via gradient descent guided by external attribute models such as pose or image quality. The model is trained on 50,000 identities from the WebFace4M dataset \cite{zhu2021webface260m}, and the authors release multiple subsets with varying numbers of generated identities.

\begin{figure*}[!h] 
               \centering
               \includegraphics[width=0.99\textwidth]{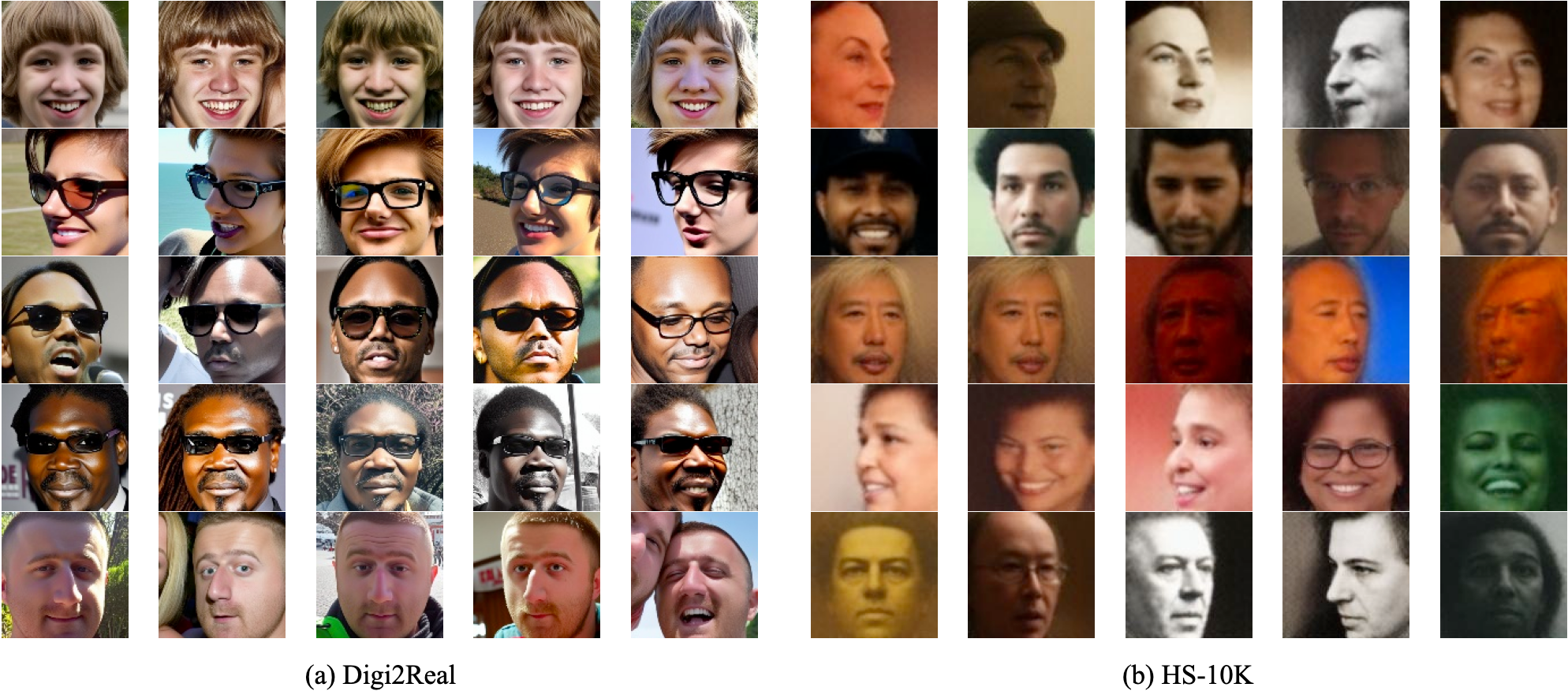}
               \caption{ Image shows intra class variations from (a) Digi2Real dataset \cite{george2025digi2real} and HS-10K \cite{wu2024vec2face}. Each row shows images from a unique identity and each column shows the intra-class variations in the identity in the dataset. }
               \label{fig:dataset_intraclass} 
\end{figure*}

\begin{table*}[htb]
\centering
\begin{tabular}{cccccccc}
\hline
\textbf{Dataset} & \textbf{No.of IDs} & \textbf{LFW} & \textbf{CFP-FP} & \textbf{AgeDB-30} & \textbf{CALFW} & \textbf{CPLFW}  \\
\hline
HS-10K \cite{wu2024vec2face} & 10K   & 98.60$\pm$0.3 & 88.87$\pm$1.9 & \textbf{92.78$\pm$0.9} & \textbf{93.67$\pm$1.1} & 85.68$\pm$1.9\\
Digi2Real-10K \cite{george2025digi2real}  & 10K   & 98.30$\pm$0.8 & 87.03$\pm$1.4 & 81.73$\pm$1.8 & 86.80$\pm$1.5 & 81.77$\pm$2.6  \\ \hline
HS-5K+Digi2Real-5K &10K  & \textbf{98.92$\pm$0.3} & \textbf{91.26$\pm$1.83} & 89.42$\pm$1.8 & 91.93$\pm$1.3 & \textbf{86.30$\pm$2.1} \\ \hline
HS-10K+Digi2Real-10K &20K & \textbf{99.13 $\pm$0.3} & \textbf{92.22 $\pm$1.12}& 91.20 $\pm$1.2& 93.17 $\pm$1.0& \textbf{86.88 $\pm$2.0} \\
\hline
\end{tabular}
\caption{Performance of different dataset combinations on standard face recognition benchmarks (10K identities).}
\label{tab:res10k}
\end{table*}

\begin{table*}[h!]
\centering
\begin{tabular}{cccccccc}
\hline
\textbf{Dataset} & \textbf{No.of IDs} & \textbf{LFW} & \textbf{CFP-FP} & \textbf{AgeDB-30} & \textbf{CALFW} & \textbf{CPLFW}  \\
\hline
HS-20K \cite{wu2024vec2face} &20K  &98.80 ± 0.28 & 89.10 ± 1.69 & \textbf{93.82 ± 1.35} & \textbf{93.85 ± 1.15} & 86.23 ± 1.61 \\
Digi2Real-20K \cite{george2025digi2real} & 20K & 98.65 ± 0.57 & 88.66 ± 1.90 & 84.93 ± 1.90 & 87.92 ± 1.46 & 81.88 ± 2.75 \\ \hline
HS-10K+Digi2Real-10K & 20K & \textbf{99.13 ± 0.28} & \textbf{92.23 ± 1.12} & 91.20 ± 1.26 & 93.17 ± 1.09 & \textbf{86.88 ± 2.01} \\ \hline
HS-20K+Digi2Real-20K &40K & \textbf{99.22 ± 0.39} & \textbf{92.07 ± 1.67} & 91.63 ± 1.34 & 93.08 ± 1.16 & \textbf{86.92 ± 1.59} \\
\hline
\end{tabular}
\caption{Performance of different dataset combinations on standard face recognition benchmarks (20K identities).}
\label{tab:res20k}
\end{table*}

\subsection{Combining the datasets} In this work, we combine subsets of two synthetic datasets, Digi2Real (5K, 10K, 20K) and Vec2Face (HS-5K, 10K, 20K), which are generated using fundamentally different pipelines: one based on graphics rendering augmented with generative refinement and the other on latent space sampling guided with external attribute models. This combination aims to leverage the complementary strengths of both approaches: the identity consistency and controllability offered by graphics-based synthesis, and attribute diversity enabled by generative methods. By merging datasets from distinct sources (Fig. \ref{fig:framework}), we seek to reduce generator-specific artifacts and increase overall data diversity, improving model generalization. Figure \ref{fig:dataset_intraclass} illustrates the intra-class variations present in the two datasets. Specifically, the quality and diversity of the samples differ significantly between them.

To ensure compatibility when combining the datasets, it is essential to verify that the identities in each dataset do not overlap. For this purpose, we visualize the distribution of identity embeddings extracted using a pretrained EdgeFace model \cite{george2024edgeface}. As shown in Figure~\ref{fig:edgeface}, the images from different datasets form different clusters, confirming that there is no overlap among them. Additionally, we examine the distribution of the datasets in the CLIP embedding space \cite{radford2021learning}, as illustrated in Figure~\ref{fig:clip}. The clear separation of samples in this space further demonstrates a significant distribution shift across the datasets.

The details of our data set combination strategy, as well as the experimental setup and evaluation protocols, are presented in the following section. 

\section{Experiments}
\label{sec:exp}

This section outlines the data used in the experiments and presents the benchmarking results.

\subsection{Benchmarking Datasets} For benchmarking the performance of the approach, we perform evaluations on several standard face recognition benchmarks, including Labeled Faces in the Wild (LFW) \cite{huang2008labeled}, Cross-Age LFW (CA-LFW) \cite{zheng2017cross}, Cross-Pose LFW (CP-LFW) \cite{zheng2018cross}, Celebrities in Frontal-Profile (CFP-FP) \cite{sengupta2016frontal}, and AgeDB-30 \cite{moschoglou2017agedb}.

\subsection{Experiments and Discussions}

For our experiments, we used the 10K-identity subset of the Vec2Face dataset \cite{wu2024vec2face}, referred to as HS-10K. Similarly, we selected a 10K-identity subset from the Digi2Real-20K dataset \cite{george2025digi2real}, denoted as Digi2Real-10K. To evaluate the effect of combining datasets, we also created subsets containing 5K identities each from Vec2Face and Digi2Real, referred to as HS-5K and Digi2Real-5K, respectively. Two combined datasets were made from these individual subsets HS-5K+Digi2Real-5K  which combines HS-5K and Digi2Real-5K , and HS-10K+Digi2Real-10K which combines HS-10K and Digi2Real-10K. Similarly, we created HS-20K+Digi2Real-20K which combines HS-20K and Digi2Real-20K.

All models were trained using the IResNet50 architecture with ArcFace loss \cite{deng2019arcface}. Training was conducted for 26 epochs with an initial learning rate of 0.1, which was reduced at epochs 12, 20, and 24 according to a step-wise learning rate schedule. The batch size used was 128 with a weight decay of 5E-4. The model was trained using SGD optimizer. The final model checkpoint was used for evaluation.

We trained models on HS-10K, Digi2Real-10K, and the two combined variants: HS-5K+Digi2Real-5K and HS-10K+Digi2Real-10K. The results are presented in Table \ref{tab:res10k}. In three out of five benchmark cases, the combined dataset HS-5K+Digi2Real-5K, despite having the same total number of identities as each individual dataset outperformed both of its constituent datasets. Further improvements were observed with the larger combined set, HS-10K+Digi2Real-10K, which includes all available identities. However, performance degradation occurred in two benchmarks, likely due to significant disparities between the original datasets. We did further tests on the 20K sets and the results in Table \ref{tab:res20k} show similar trends but with better overall performance. These experiments suggest that combining synthetic datasets, when balanced effectively, can lead to improved face recognition performance.

\subsection{DataCV ICCV Challenge Submission}

The DataCV Challenge, in conjunction with the ICCV 2025, focuses on the evaluation of synthetic face recognition datasets. The goal of the competition is to develop methods for generating synthetic face recognition training datasets that match or surpass the accuracy of models trained on real identities, with dataset quality assessed based on the performance of face recognition models trained on the generated data. Specifically, the competition included three categories based on the number of identities in the training dataset: 10K, 20K, and 100K, with a maximum of 50 images per identity. The organizers provided a standardized training pipeline that included an IResNet50 architecture with fixed hyperparameters. A fixed test set was also provided, and rankings were determined by evaluating model performance on this set. The final competition ranking was based on the 10K category, for which our submission used a combined dataset of 5K identities from Digi2Real and 5K from HS-5K. Our submission achieved second place in the 10K track during the test phase of the competition.

\section{Conclusions}

In this work, we explore how the combination of two synthetic datasets, each generated using distinct pipelines can enhance face recognition performance. Our experiments demonstrate that even on the same identity scale, merging datasets from different generation methods can lead to improved performance. Although the combination strategy used here is relatively simple, future research could investigate more targeted sample selection to maximize diversity and effectiveness in the combined dataset. Further it would be possible to combine multiple datasets together with an effective sampling strategy, which can increase the diversity of the combined dataset and could make it more robust. To support reproducibility, we will publicly release the training protocols. We hope these findings encourage further research on effective strategies for combining synthetic datasets to address the performance gap with models trained on real-world data.

\section{Acknowledgements}
This research was funded by the European Union project CarMen (Grant Agreement No. 101168325).
{
    \small
    \bibliographystyle{ieeenat_fullname}
    \bibliography{main}
}

\end{document}